\documentclass{article}

\usepackage[preprint,nonatbib]{neurips_2020}

\usepackage[utf8]{inputenc} 
\usepackage[T1]{fontenc}    
\usepackage{hyperref}       
\usepackage{url}            
\usepackage{booktabs}       
\usepackage{amsfonts}       
\usepackage{nicefrac}       
\usepackage{microtype}      

\usepackage{latexsym}

\usepackage{graphicx}
\usepackage{amsmath}
\usepackage{amssymb}
\usepackage{multirow}
\usepackage{setspace}
\usepackage{CJK}
\usepackage{enumitem}
\DeclareMathOperator*{\argmax}{argmax}

\usepackage{diagbox}
\usepackage{wrapfig}
\usepackage{xcolor}
\usepackage[]{todonotes}

\title{Defense against Adversarial Attacks in NLP via Dirichlet Neighborhood Ensemble}

\author{%
  Yi Zhou\textsuperscript{1}  \quad
  Xiaoqing Zheng\textsuperscript{1,*}  \quad
  Cho-Jui Hsieh\textsuperscript{2} \quad
  Kai-wei Chang\textsuperscript{2} \quad
  Xuanjing Huang\textsuperscript{1} \\
  \textsuperscript{1}Fudan University, Shanghai 201203\\
  \textsuperscript{2}University of California, Los Angeles, Los Angeles CA 90095 \\
  \texttt{yizhou17@fudan.edu.cn}, \quad
  \texttt{zhengxq@fudan.edu.cn}, \\
  \texttt{chohsieh@cs.ucla.edu}, \quad
  \texttt{kw@kwchang.net}, \quad
  \texttt{xjhuang@fudan.edu.cn}\\
}

\begin{document}

\maketitle

\begin{abstract}

Despite neural networks have achieved prominent performance on many natural language processing (NLP) tasks, they are vulnerable to adversarial examples. In this paper, we propose Dirichlet Neighborhood Ensemble (DNE), a randomized smoothing method for training a robust model to defense substitution-based attacks. During training, DNE forms virtual sentences by sampling embedding vectors for each word in an input sentence from a convex hull spanned by the word and its synonyms, and it augments them with the training data. In such a way, the model is robust to adversarial attacks while maintaining the performance on the original clean data. 
DNE is agnostic to the network architectures and scales to large models for NLP applications.
We demonstrate through extensive experimentation that our method consistently outperforms recently proposed defense methods by a significant margin across different network architectures and multiple data sets.







\end{abstract}

\section{Introduction}

Deep neural networks are powerful but vulnerable to adversarial examples that are intentionally crafted to fool the models.
To address this issue, adversarial attacks and defenses against these attacks have drawn significant attention in recent years \cite{arxiv-13:Szegedy,iclr-15:Goodfellow,cvpr-16:Moosavi-Dezfooli,sp-16:Papernot,sp-17:Carlini,iclr-17:Miyato,iclr-18:Madry,cvpr-18:Eykholt,nips-18:Wong,icml-19:Cohen,iclr-20:Zhang}.
In the context of natural language processing (NLP), generating adversarial examples for texts has shown to be a more challenging task than for images and audios due to their discrete nature. However, several recent studies have demonstrated the vulnerability of deep neural networks in NLP tasks, including reading comprehension \cite{emnlp-17:Jia}, text classification \cite{arxiv-17:Samanta,arxiv-17:Wong,ijcai-18:Liang, emnlp-18:Alzantot}, machine translation \cite{iclr-18:Zhao,acl-18:Ebrahimi,arxiv-18:Cheng}, dialogue systems \cite{naacl-19:Cheng}, and dependency parsing \cite{acl-20:Zheng}. 
These methods often attack an NLP model by replacing, scrambling, and erasing characters or words under certain semantic and syntactic constraints.
In particular, most of them construct adversarial examples by substituting words with their synonyms in an input text to maximally increase the prediction error while maintaining the fluency and naturalness of the adversarial examples. 
In this study, we consider such a word substitution-based threat model and discuss the strategy to defend such an attack. 

The goal of adversarial defenses is to learn a model that is capable of achieving high test accuracy on both clean and adversarial examples.
Adversarial training is one of the most successful defense methods for NLP models~\cite{iclr-17:Miyato,ijcai-18:Sato,iclr-19:Zhu}. 
During the training time, they replace a word by one of its synonyms that approximately maximizes the prediction loss.
By training on these adversarial examples, the model is robust to such perturbations.
However, the relative positions between word vectors of a word and its synonyms change dynamically during training as their embeddings are updated independently. 
The point-wise guarantee becomes insufficient, and the resulting models have shown to be vulnerable to strong attacks \cite{emnlp-18:Alzantot}.
On the other hand, recently several certified defense methods have been proposed to improve over adversarial training~\cite{emnlp-19:Jia,emnlp-19:Huang} by 
certifying the performance within the convex hull formed by the embeddings of a word and its synonyms.  However, due to the difficulty of propagating convex hull through deep neural networks, they compute a very loose outer bound using Interval Bound Propagation (IBP). 
As the result, the convex hall may contain irrelevant words and lead to a significant performance drop on the clean data. 

In this paper, 
we propose {\bf Dirichlet Neighborhood Ensemble (DNE)} to create virtual sentences by \emph{mixing} the embedding of the original word in the input sentence with its synonyms. By training on these virtual sentences, the model can enhance the robustness against word substitution-based perturbations. 
Specifically, our method samples an embedding vector in the convex hull formed by a word and its synonyms to ensure the robustness within such a region. In contrast to IBP, our approach better represents the subspace of the synonyms when creating the virtual sentences.
A gradient-guided optimizer is then applied to search for more valuable adversarial points within the convex hull, and the framework can be extended to higher-order neighbors (synonyms) to further boost the robustness. 
In the inference time, the same Dirichlet sampling technique is used again and the prediction scores on the virtual sentences are ensembled to get a robust output. 

Through extensive experiments with various model architectures (bag-of-words, CNN, LSTM, and attention-based) on multiple data sets, we show that DNE consistently achieves better performance on clean and adversarial samples compared with existing defense methods. 
By conducting a detailed analysis, we found that DNE enables the embeddings of a set of similar words to be updated together in a coordinated way. In contrast, prior approaches either fix the word vectors during training (e.g., in the certified defenses) or update individual word vectors independently (e.g., in the adversarial training). We believe this is the key property why DNE leads to a more robust NLP model.
Furthermore, unlike the certified defenses, the proposed method is easy to implement and can be integrated into any existing neural networks including the ones with large architecture such as BERT \cite{naacl-18:Devlin}.


\section{Related Work}


In the text domain, adversarial training so far is one of the most successful defenses according to many recent studies \cite{iclr-17:Miyato,ijcai-18:Sato,iclr-19:Zhu}.
A family of fast-gradient sign methods (FGSM) was introduced by Goodfellow et al. \cite{iclr-15:Goodfellow} to generate adversarial examples in the image domain, and they showed that the robustness and generalization of machine learning models can be improved by including high-quality adversaries in the training data.
Miyato et al. \cite{iclr-17:Miyato} proposed a FGSM-like adversarial training method to the text domain by applying perturbations to the word embeddings rather than to the original input itself.
Sato et al. \cite{ijcai-18:Sato} extended the work of \cite{iclr-17:Miyato} to improve the interpretability by constraining the directions of perturbations toward the existing words in the word embedding space. Given a word, such direction is calculated by the weighted sum of unit vectors from the word to its nearest neighbors. 
Barham et al. \cite{arxiv-19:Barham} presented a sparse projected gradient descent (SPGD) method to impose a sparsity constraint on perturbations by projecting them onto the directions to nearby word embeddings with the highest cosine similarities.

Zhang and Yang \cite{arxiv-18:Zhang} applied several types of noises to perturb the input word embeddings, such as Gaussian, Bernoulli, and adversarial noises, to mitigate the overfitting problem of NLP models. For the adversarial noise, the perturbation is added in the direction of maximally increasing the loss function. 
Zhu et al. \cite{iclr-19:Zhu} proposed a novel adversarial training algorithm, called FreeLB (Free Large-Batch), which adds adversarial perturbations to word embeddings and minimizes the resultant adversarial loss inside different regions around input samples.
They add norm-bounded adversarial perturbations to the embeddings of the input sentences using a gradient-based method and enlarge the batch size with diversified adversarial samples under such norm constraints. However, they focus on the effects on generalization rather than the robustness against adversarial attacks.




Although adversarial training can empirically defense the attack algorithms used during the training, the trained model often still cannot survives from another sophisticated attacks. Recently a set of certified defenses have been introduced, which provide guarantees of robustness to some specific types of attacks.
For example, Jia et al. \cite{emnlp-19:Jia} and Huang et al. \cite{emnlp-19:Huang} use a bounding technique, interval bound propagation (IBP) \cite{arxiv-18:Gowal,arxiv-18:Dvijotham}, to formally verify a model's robustness against word substitution-based perturbations. 
Shi \cite{iclr-20:Shi} proposed the first robustness verification method for transformers by IBP-style technique.
However, these defenses often lead to loose upper bounds for arbitrary networks and result in the greater cost of clean accuracy.
Furthermore, most techniques developed so far require knowledge of the architecture of the machine learning models and still remain hard to scale to complex prediction pipelines.

In the image domain, randomization has been shown to overcome many of these obstacles in IBP-based defense. Empirically, 
Xie et al. \cite{xie2017mitigating} showed that random resizing and padding in the input domain can improve the robustness. Liu et al. \cite{liu2018towards} proposed to add Gaussian noise in both input layer and intermediate layers of CNN in both training and inference time to improve the robustness. Lecuyer et al. \cite{lecuyer2019certified} provided a certified guarantee of this method, and later on the bound is significantly improved in \cite{icml-19:Cohen}. The resulting algorithm, called random smoothing, has become widely used in certifiying $\ell_2$ robustness for image classifiers. 
To the best of our knowledge, these random smoothing methods have not been used in NLP models, and the main reason is that the adversarial examples in texts are usually generated by word substitution-based perturbations instead of small $\ell_p$ norm, and as shown in our experiments, randomly perturbing a word to its synonyms performs poorly in practice. 
The proposed algorithm can be viewed as a kind of randomized defense on NLP models, where our main contribution is to show that it is important to ensure the model works well in a region within the convex hull formed by the embeddings of a word and its synonyms instead of only ensuring model is good under discrete perturbation. Furthermore, we show the method can be combined with adversarial training to further boost the empirical robust accuracy. 

\section{Method} \label{sec:method}

Let $f$ be a base classifier which maps an input sentence $x \in \mathcal{X}$ to a class label $y \in \mathcal{Y}$. We consider the setting where for each word $x_i$ in the sentence $x$, we are given a set of its synonyms $\mathcal{S}(x_i)$ including $x_i$ itself, where we know replacing $x_i$ by any of $\mathcal{S}(x_i)$ is unlikely to change the semantic meaning of the sentence\footnote{Follow \cite{emnlp-19:Jia},
we base our sets of allowed word substitutions $S(x_i)$ on the substitutions proposed by Alzantot et al \cite{emnlp-18:Alzantot}.
They compute the eight nearest neighbors of the selected word according to the distance in the GloVe embedding space \cite{emnlp-14:Pennington}, and then use the counter-fitting method \cite{naacl-16:Mrksic} to post-process the adversary's GloVe vectors to ensure that the nearest neighbors are synonyms.}.
We relax the set of discrete points (a word and its synonyms) to a convex hull spanned by the word embeddings of all these points, denoted by $\mathcal{C}(x_i)$.
We assume any perturbation within this convex hull will keep the semantic meaning unchanged, and define a smoothed classifier $g(x)$ based on random sampling within the convex hull. In the training time, the base classifier is trained with ``virtual'' data augmentation in the embedding space, where each $x_i$ is replaced by a point in the convex hull containing $\mathcal{C}(x_i)$ by the proposed sampling algorithm described below. 
A novel adversarial training algorithm is also used to enable NLP models to defense against the strong attacks that search for the worst-case over all combinations of word substitutions. 
In the inference time, a similar sampling strategy is conducted and a CBW-D ensemble algorithm \cite{cvpr-19:Dubey} is used to compute the final prediction. 

Note that it is impossible to exactly calculate the probabilities with which $f$ classifies $x$ as each class, so we use a Monte Carlo algorithm for evaluating $g(x)$. 
As an illustration in Fig. 1 (a), for an input sentence $x$, we draw $k$ samples of $\hat{x}$ by running $k$ noise-corrupted copies of $x$ through the base classifier $f(\hat{x})$, where $\hat{x}$ is generated by replacing the embedding of every word $x_j$ with a point randomly sampled with the Dirichlet distribution from $\mathcal{C}(x_j)$ (the pentagon with yellow dashed borders).  
If the class $y$ appeared with maximal weight in the categorical distribution $\hat{x}$, the smoothed classifier $g(x)$ returns $y$.
In the following, we introduce each component of the proposed algorithm.


\subsection{Dirichlet Neighborhood Sampling} \label{sec:smooth}
The random perturbations of $x$ are combinatorial in nature, and thus training the base classifier $f$ that consistently labels any perturbation of $x$ as $y$ requires checking an exponential number of predictions.
To better reflect those discrete word substitution-based perturbations, we sample the points from a convex hull using the Dirichlet distribution.
This allows us to control how far we can expect the points are from any vertex of the convex hull.
If a sampled point is very close to a vertex (i.e., a word), it simulates a word substitution-based perturbation in which the vertex is chosen to replace the original one.
Any point sampled from $\mathcal{C}(x_i)$ is a convex combination of the embeddings of $\mathcal{S}(x_i)$: 
\begin{equation} \label{eq:representation}
    \nu(x_i) = \sum_{x_j \in \mathcal{S}(x_i)} \beta_j \cdot \boldsymbol{x_j}, 
\end{equation}
where $\beta_j \geq 0$, $\Sigma_{j} \beta_j = 1$, and $\boldsymbol{x_j}$ (in bold type) denotes the embedding of $x_j$.
A vector $\boldsymbol{\beta}$ contains the weights drawn from the Dirichlet distribution as follows:
\begin{equation} \label{eq:sample}
    \beta_1, \dots, \beta_m \sim \text{Dir}(\alpha_1, \dots, \alpha_m), 
\end{equation}
where $m$ is the size of $\mathcal{S}(x_i)$, and the Dirichlet distribution is parameterized by a vector of $\boldsymbol{\alpha}$ used to control the degree in which the words in $\mathcal{S}(x_i)$ contribute to generate the vector $\nu(x_i)$.
There are two extreme cases. If $\boldsymbol{\alpha} = \boldsymbol{0}$, only one of $\mathcal{S}(x_i)$ is sampled to replace $x_i$; If $\boldsymbol{\alpha} = \boldsymbol{\infty}$, the result $\nu(x_i)$ is equal to the average of the embeddings of all of the words in $\mathcal{S}(x_i)$.

\begin{figure}
  \centering
  \includegraphics[width = 10.5cm]{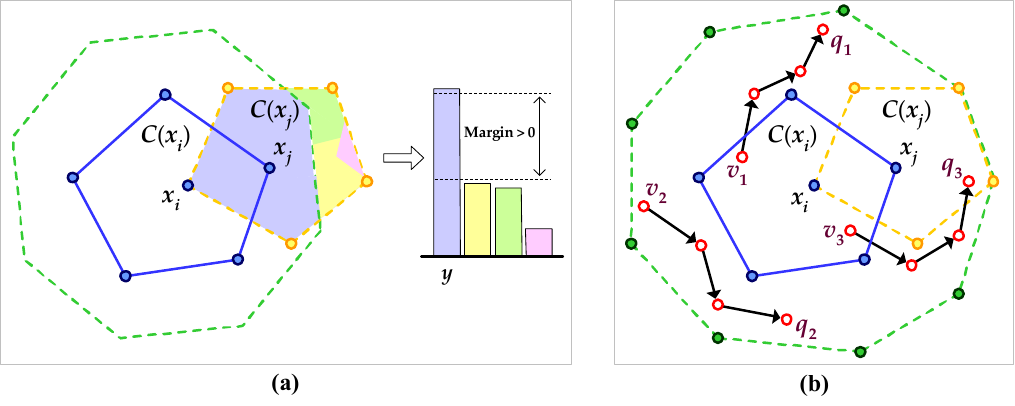}
  \label{fig:hull}
  \caption{Consider a word (sentence of length one) $x_i$ and its convex hull $\mathcal{C}(x_i)$ (projected to 2D for illustration) spanned by the set of its synonyms (blue circles).
  We assume that an adversary replaces $x_i$ with one of its synonyms $x_j$.
  \textbf{(a)} Evaluating the smoothed classifier at the input $x_j$. The decision regions of the base classifier $f$ are drawn in different colors. If we expand $\mathcal{C}(x_i)$ to the polygon with green dashed borders when training the base classifier $f$, the size of the intersection of this polygon and $\mathcal{C}(x_j)$ is large enough to ensure that the smoothed classifier $g$ labels $x_j$ as $f(x_i)$.
  Here, $g(x_j)$ is ``blue.''
  \textbf{(b)} An example convex hull used to train the base classifier. 
  Since the size of the intersection of $\mathcal{C}(x_i)$ and $\mathcal{C}(x_j)$ is small, we expand $\mathcal{C}(x_i)$ to the convex hull spanned by $x_i$'s neighbors and ``neighbors of neighbors'' in their embedding space when training the base classifier $f$. Starting from three points $\boldsymbol{v_1}$, $\boldsymbol{v_2}$ and $\boldsymbol{v_3}$ sampled from the expanded convex hull (the largest polygon with green dashed borders), $\boldsymbol{q_1}$, $\boldsymbol{q_2}$ and $\boldsymbol{q_3}$ are the local ``worst-case'' points found by searing over the entire convex hull with the gradient-guided optimization method.}
\end{figure}

\subsection{Training the Base Classifier with Two-Hop Neighbors}

For the smoothed classifier $g$ to classify an adversarial example of $x$ correctly and robustly, $f$ needs to consistently classify $\hat{x}$ as the gold label of $x$. 
Therefore, we train the base classifier with virtual data augmentation $\hat{x}$ for each training example $x$.
In Fig. 1 (b), we illustrate the process by considering a sentence with one word $x_i$ and the set of its synonyms (shown as blue circles).
The input perturbations span a convex hull of $\mathcal{C}(x_i)$ around the word $x_i$ (the pentagon with blue borders, projected to 2D here). 
Assuming that the word $x_i$ is replaced with $x_j$ by an adversary, noise-corrupted samples will be drawn from $\mathcal{C}(x_j)$ (the pentagon with yellow dashed borders) instead of $\mathcal{C}(x_i)$.
If the size of the intersection of $\mathcal{C}(x_i)$ and $\mathcal{C}(x_j)$ is small, we cannot expect $f$ will consistently classify $x_j$ as the same label as $x_i$.
Therefore, we expand $\mathcal{C}(x_i)$ to the convex hull spanned by the word embeddings of the union of $\mathcal{S}(x_i)$ and all of $\mathcal{S}(x_j), x_j \in \mathcal{S}(x_i)$, namely $x_i$'s 1-hop neighbors and 2-hop neighbors in their embedding space, denoted by $\mathcal{B}(x_i)$.

Such expansions will slightly hurt the performance on the clean data.
Recall that different values of $\boldsymbol{\alpha}$ can be used to control the degree in which the 1-hop and 2-hop neighbors to contribute to generate $\widetilde{x}$.
In our implementation, we let the expected weights of the 2-hop neighbors are less than one-half of those of the 1-hop neighbors when computing $\widetilde{x}$ as Eq. \eqref{eq:representation} to reduce the impact on the clean accuracy.
We use $\mathcal{B}(x_i)$ to denote the expanded convex hull of $\mathcal{C}(x_i)$, and $\widetilde{x}$ to a virtual example created by replacing the embedding of every word $x_i$ in an input sentence $x$ with a point randomly sampled from $\mathcal{B}(x_i)$ by the Dirichlet distribution.

The base classifier is trained by minimizing the cross-entropy error with virtual data augmentation by the gradient descent.
We assume the base classifier takes form $f(x) = \argmax_{c \in \mathcal{Y}} s_c(x)$, where each $s_c(x)$ is the scoring function for the class $c$. That is, the outputs of the neural networks before the softmax layer. Our objective is to maximize the sum of the log-probabilities that $f$ will classify each $\widetilde{x}$ as the label of $x$.
Let $\mathcal{D}$ be a training set of $n$ instances, and each of them is a pair of $(x, y)$:
\begin{equation} \label{eq:objective}
  \sum_{\forall (x, y) \in \mathcal{D}}
  \log \mathbb{P}_{\widetilde{x}}(f(\widetilde{x}) = y) =
  \sum_{\forall (x, y) \in \mathcal{D}}
  \log \mathbb{E}_{\widetilde{x}} \mathbf{1} 
  \left [ \argmax_{c \in \mathcal{Y}} s_{c}(\widetilde{x}) = y \right ], 
\end{equation}
where $\widetilde{x}$ is a virtual example randomly created for an input example $x$. The softmax function can be viewed as a continuous, differentiable approximation of argmax:
\begin{equation} 
  \mathbf{1} 
  \left [ \argmax_{c \in \mathcal{Y}} s_{c}(\widetilde{x}) = y \right ]
  \approx \frac{\exp(s_{y}(\widetilde{x}))}{\sum_{c \in \mathcal{Y}}\exp(s_{c}(\widetilde{x}))}.
\end{equation}
By the concavity of log and Jensen's inequality, the objective is approximately lower-bounded by:
\begin{equation} \label{eq:loss}
  \sum_{\forall (x, y) \in \mathcal{D}}
  \mathbb{E}_{\widetilde{x}} \left [ \log \frac{\exp(s_{y}(\widetilde{x}))}{\sum_{c \in \mathcal{Y}}\exp(s_{c}(\widetilde{x}))} \right ].
\end{equation} 
This is the negative cross-entropy loss with virtual data augmentation.
Maximizing Eq. \eqref{eq:loss} approximately maximizes Eq. \eqref{eq:objective}.

Since the virtual data point defined in Eq. \eqref{eq:representation} is a linear combination of embeddings of $\mathcal{S}(x_i)$, the back-propagation will propagate the gradient to all these embeddings with nonzero coefficients, thus allowing updating all these embeddings together in a coordinated way when performing parameter updates. 
As illustrated in Fig. 1, the whole green convex hull will be shifted together at each iteration.
In contrast, traditional adversarial training only updates the embedding of one synonym (a vertex of the convex hull), which will distort the relative position of those embeddings and thus become slower and less stable. 
It is probably why the word embeddings are fixed during training in the certified defenses \cite{emnlp-19:Huang,emnlp-19:Jia}.
Even though the word embeddings can be pre-trained, holding embeddings fixed makes them impossible to be fine-tuned for the tasks of interest, which may hurt the performance.

\subsection{Adversarial Training} \label{sec:adversarail}

To promote higher robustness and invariance to any region within the convex hull, we further propose to combine Dirichlet sampling with adversarial training to better explore different regions inside the convex hull $\mathcal{B}(x_i)$.
Any point sampled from $\mathcal{B}(x_i)$ is represented as the convex combination of the embeddings of its vertices, which ensures that a series of points keep stay inside of the same $\mathcal{B}(x_i)$ while searching for the worst-case over the entire convex hull by any optimization method.

Assuming that a virtual example $\widetilde{x}$ is generated for an input text $x$, we search for the next adversarial example to maximize the model’s prediction error by updating every vector of weights $\boldsymbol{\beta} = \exp (\boldsymbol{\eta})$ by the following formula, each of them is used to represent a point sampled from $\mathcal{B}(x_i)$ as Eq. \eqref{eq:representation}:
\begin{equation} \label{eq:update}
  \boldsymbol{\eta} \leftarrow \boldsymbol{\eta} - \epsilon \left\Vert \frac{\partial \log p(\widetilde{x}, y)}{\partial \boldsymbol{\eta}} \right\Vert_2, \,\,
  p(\widetilde{x}, y) = \frac{\exp(s_{y}(\widetilde{x}))}{\sum_{c \in \mathcal{Y}}\exp(s_{c}(\widetilde{x}))}, 
\end{equation}
where $\epsilon$ is the step size. In order to ensure that the updated $\boldsymbol{\beta}$ still satisfy $\beta_j \geq 0$ and $\Sigma_{j} \beta_j = 1$, we sequentially apply logarithmic and softmax functions to $\boldsymbol{\beta}$ after it is randomly drawn from $\text{Dir}(\boldsymbol{\alpha})$. Note that $\text{softmax}(\log(\boldsymbol{\beta})) = \boldsymbol{\beta}$, and $\boldsymbol{\eta}$ will be updated instead of $\boldsymbol{\beta}$ in our implementation. 
By updating $\boldsymbol{\eta}$ only, the representation defined in Eq. \eqref{eq:representation} also ensures that a series of points keep stay inside of the same convex hull while searching for the worst-case over $\mathcal{B}(x_i)$ by gradient-guided optimization methods.

As illustrated in Fig. 1 (b), we apply this update multiple times with small step size (arrow-linked red circles represent data points generated after each update by adding gradient-guided perturbations to their preceding ones). When training the base classifier $f$, we add all of the virtual examples generated at every search step (i.e., all of the points indicated by the red circles in Fig. 1 (b)) into the training set to better explore different regions around $x$. 

\subsection{Ensemble Method} \label{sec:ensemble}
As mentioned above, if the base classifier $f$ is a neural network, it is impossible to exactly calculate the probabilities with which $f$ classifies $x$ as each class.
Following randomized defense in computer vision~\cite{liu2018towards,lecuyer2019certified,icml-19:Cohen}, we use a Monte Carlo algorithm for evaluating $g(x)$.
Given an input sentence $x$, we draw $k$ Monte Carlo samples of $\hat{x}$ by running $k$ noise-corrupted copies of $x$ through the base classifier $f(\hat{x})$, where each $\hat{x}$ is created by replacing the embedding of every word $x_i$ in the sentence $x$ with a point randomly sampled with the Dirichlet distribution from $\mathcal{C}(x_i)$ (not from the expanded convex hull $\mathcal{B}(x_i)$ when testing).

We combine predictions by taking a weighted average of the softmax probability vectors of all the randomly created $\hat{x}$, and take the argmax of this average vector as the final prediction. We choose to use CBW-D \cite{cvpr-19:Dubey} to compute those weights. The idea behind it is to give more weights the predictions that have more confidence in their results. CBW-D calculates the weights $w$ as a function of the differences between the maximum value of the softmax distribution and the other values as follows:
\begin{equation} \label{combine}
  w = \sum_{c \in \mathcal{Y}, c \neq y} (p(\hat{x}, y) - p(\hat{x}, c))^r, 
\end{equation}
where $y$ is the class having the maximum probability in a prediction, $r$ is a hyperparameter tuned using cross-validation in preliminary experiments.




\section{Experiments}
\label{experiments}

We conducted experiments on multiple data sets for text classification and natural language inference tasks. 
Various model architectures (bag-of-words, CNN, LSTM, and attention-based) were used to evaluate our Dirichlet Neighborhood Ensemble (DNE) and other defense methods under two recently proposed attacks \cite{emnlp-18:Alzantot,acl-19:Ren}.
Ren et al. \cite{acl-19:Ren} described a greedy algorithm, called Probability Weighted Word Saliency (PWWS), for text adversarial attack based on word substitutions with synonyms. 
The word replacement order is determined by taking both word saliency and prediction probability into account. 
Alzantot et al. \cite{emnlp-18:Alzantot} developed a generic algorithm-based attack, denoted by GA, to generate semantically and syntactically similar adversarial examples. They also use a language model (LM) \cite{arxiv-18:Chelba} to rule out candidate substitute words that do not fit within the context.
However, unlike PWWS, ruling out some candidates by the LM will greatly reduce the number of candidate substitute words ($65\%$ off in average).
For fair comparison, we report the robust accuracy under GA attack both with and without using the LM.
For each date set, we measure accuracy on perturbations found by the two attacks (PWWS and GA) on $1000$ randomly selected test examples.

We primarily compare with the adversarial training (ADV) \cite{naacl-19:Michel} and the interval bound propagation (IBP) based methods \cite{emnlp-19:Huang,emnlp-19:Jia}.
The former can improve model's robustness without suffering much drop on the clean input data by adding adversarial examples in the training stage.
The latter was shown to be more robust to word substitution-based perturbations than ones trained with data augmentation. 
To demonstrate that mixing the embedding of the original word with its synonyms performs better than naively replacing the word with its synonyms, we developed a strong baseline, denoted by RAN. 
The models trained by RAN will take as inputs the corrupted copy of each input sentence, in which every word of the sentence is randomly replaced with one of its synonyms. In the inference time, the same random replacement is used and the prediction scores are ensembled to get an output. RAN can be viewed as a naive way to apply random smoothing to NLP models.

\subsection{Text Classification}

We experimented on two text classification data sets: Internet Movie Database (IMDB) \cite{acl-11:Maas}
and AG News corpus (AGNEWS) \cite{nips-15:Zhang}.
IMDB has $50,000$ movie reviews for binary (positive or negative) sentiment classification, and AGNEWS consists of about $30,000$ news articles pertaining to four categories.
We implemented three models for these text classification tasks. 
The bag-of-words model (BOW) averages the word embeddings for each word in the input, then passes this through a one-layer feedforward network with $100$-dimensional hidden state to get a final logit. The other two models are similar, except they run either a CNN or a two-layer LSTM on the word embeddings. All models are trained on cross entropy loss, and their hyper-parameters are tuned on the validation set.
Implementation details are provided in Appendix A.1.

\begin{table*} [htbp] \small

\caption{\label{tb:imdb} Text classification on IMDB dataset.}
\begin{center}
\setlength{\tabcolsep}{1.0mm}
\begin{tabular}{l|cccc|cccc|cccc}

\hline
\hline

\multirow{2}{*}{\bf IMDB} & \multicolumn{4}{c|}{\bf BOW} & \multicolumn{4}{c|}{\bf CNN} & \multicolumn{4}{c}{\bf LSTM} \\ \cline{2-13}
& {\bf CLN} & {\bf PWWS} & {\bf GA-LM} & {\bf GA}
& {\bf CLN} & {\bf PWWS} & {\bf GA-LM} & {\bf GA}
& {\bf CLN} & {\bf PWWS} & {\bf GA-LM} & {\bf GA} \\ 

\hline

{\bf ORIG} & $\bf 89.9$ & $4.1$ & $1.2$ & $0.4$ & $\bf 90.2$ & $18.1$ & $4.2$ & $2.0$ & $\bf 89.8$ & $0.2$ & $2.1$ & $0.0$ \\

{\bf ADV} & $86.4$ & $77.4$ & $80.0$ & $\bf 77.2$ & $87.0$ & $72.1$ & $76.0$ & $72.0$ & $85.6$ & $35.4$ & $56.6$ & $32.0$ \\

{\bf IBP} & $79.6$ & $75.4$ & $70.5$ & $66.9$ & $79.6$ & $76.3$ & $75.0$ & $70.9$ & $76.8$ & $72.2$ & $64.7$ & $64.3$ \\

{\bf RAN} & $89.7$ & $39.8$ & $36.2$ & $9.2$ & $88.9$ & $27.2$ & $36.5$ & $13.3$ & $89.7$ & $37.7$ & $40.5$ & $8.1$ \\

{\bf DNE} & $86.6$ & $\bf 82.0$ & $\bf 80.5$ & $\bf 77.2$ & $87.9$ & $\bf 82.3$ & $\bf 81.2$ & $\bf 76.5$ & $88.2$ & $\bf 82.3$ & $\bf 80.5$ & $\bf 77.2$ \\

\hline
\hline

\end{tabular}
\end{center}

\end{table*}

\begin{table*} [htbp] \small

\caption{\label{tb:agnews} Text classification on AGNEWS dataset.}

\begin{center}
\setlength{\tabcolsep}{1.0mm}
\begin{tabular}{l|cccc|cccc|cccc}

\hline
\hline

{\bf AG} & \multicolumn{4}{c|}{\bf BOW} & \multicolumn{4}{c|}{\bf CNN} & \multicolumn{4}{c}{\bf LSTM} \\ \cline{2-13}
{\bf NEWS} & {\bf CLN} & {\bf PWWS} & {\bf GA-LM} & {\bf GA}
& {\bf CLN} & {\bf PWWS} & {\bf GA-LM} & {\bf GA}
& {\bf CLN} & {\bf PWWS} & {\bf GA-LM} & {\bf GA} \\ 

\hline

{\bf ORIG} & $\bf 89.4$ & $49.5$ & $57.6$ & $17.2$ & $\bf 89.0$ & $35.0$ & $46.0$ & $12.1$ & $\bf 92.5$ & $46.2$ & $52.8$ & $9.8$ \\

{\bf ADV} & $88.8$ & $84.5$ & $85.7$ & $82.5$ & $88.4$ & $80.2$ & $82.5$ & $75.3$ & $92.4$ & $85.4$ & $87.1$ & $78.8$ \\

{\bf IBP} & $87.4$ & $85.1$ & $86.8$ & $81.3$ & $87.8$ & $86.2$ & $\bf 86.7$ & $82.7$ & $84.0$ & $82.3$ & $82.9$ & $77.9$ \\

{\bf RAN} & $89.0$ & $78.1$ & $75.2$ & $51.3$ & $88.7$ & $78.2$ & $74.4$ & $51.7$ & $92.1$ & $81.4$ & $81.4$ & $51.9$ \\

{\bf DNE} & $87.8$ & $\bf 86.7$ & $\bf 87.0$ & $\bf 85.9$ & $87.3$ & $\bf 85.7$ & $85.9$ & $\bf 85.2$ & $91.9$ & $\bf 90.9$ & $\bf 90.6$ & $\bf 89.5$ \\

\hline
\hline

\end{tabular}
\end{center}

\end{table*}

In Table \ref{tb:imdb}, we present both clean accuracy (CLN) and accuracy under two attack algorithms (PWWS and GA) on IMDB with three different model architectures (BOW, CNN and LSTM). 
We use GA-LM to denote the GA-based attack that rules out candidate substitute words that may not fit well with the context with the help of the LM \cite{arxiv-18:Chelba}, and ORIG to the testing and adversarial accuracy of the models trained as usual without using any defense method.

As we can see from Table \ref{tb:imdb}, DNE ($k = 16$) outperforms ADV and IBP on the clean input data, and consistently performs better than the competitors across the three different architectures under all of the attacks we consider.
For the text classification, LSTMs seem more vulnerable to adversarial attacks than BOWs and CNNs.
Under the strongest attack GA, while the accuracies of LSTMs trained by ORIG, ADV, IBP, and RAN dropped to $0.0\%$, $32\%$, $64.3\%$, and $8.1\%$ respectively, the LSTM trained by DNE still achieved $77.2\%$ accuracy.
The results on AGNEWS are reported in Table \ref{tb:agnews}, and we found the similar trends as those on IMDB.
Any model performed on AGNEWS shows to be more robust than the same one on IMDB.
It is probably because the average length of the sentences in IMDB ($255$ words in average) is much longer than that in AGNEWS ($43$ words in average).
Longer sentences allow the adversaries to apply more word substitution-based perturbations to the examples.
Generally, DNE performs better than IBP and comparable to ADV on the clean data, while it outperforms the others in all other cases with only one exception of $86.7\%$ (just $0.8\%$ difference) achieved by IBP with CNN under GA-LM attack. 
The results for both datasets show that our DNE consistently achieves better clean and robust accuracy compared with existing defenses.

\subsection{Natural Language Inference}

We conducted the experiments of natural language inference on Stanford Natural Language Inference (SNLI) \cite{emnlp-15:Bowman} corpus, which is a collection of $570,000$ English sentence pairs (a premise and a hypothesis) manually labeled for balanced classification with the labels entailment, contradiction, and neutral.
We also implemented three models for this task.
The bag-of-words model (BOW) encodes the premise and hypothesis separately by summing their word vectors, then feeds the concatenation of these encodings to a two-layer feedforward network. 
The other two models are similar, except they run either a Decomposable Attention (DecomAtt) \cite{emnlp-16:Parikh} or BERT \cite{naacl-18:Devlin} on the word embeddings to generate the sentence representations, which uses attention between the premise and hypothesis to compute richer representations of each word in both sentences.
All models are trained on cross entropy loss, and their hyper-parameters are tuned on the validation set (see Appendix A.2).

As reported in Table \ref{tb:snli}, DNE generally performs better than the others on the robust accuracy while suffering little performance drop on the clean data on SNLI. 
Although our proposed baseline RAN ($k = 16$) achieves a slightly higher accuracy (just $1\%$ difference) with BERT under PWWS attack, it's accuracy rapidly drops to $27\%$ under the more sophisticated attack GA, where DNE still yields $61.6\%$ in accuracy.
The results on SNLI show that DNE can be applied to attention-based models like DecomAtt and scale well to large architectures such as BERT.
We leave the results of IPB with BERT as unknown because there is still a question whether IBP-based method can be applied to BERT.


\begin{table*} [htbp] \small

\caption{\label{tb:snli} Natural language inference on SNLI dataset.}

\begin{center}
\setlength{\tabcolsep}{1.0mm}
\begin{tabular}{l|cccc|cccc|cccc}

\hline
\hline

\multirow{2}{*}{\bf SNLI} & \multicolumn{4}{c|}{\bf BOW} & \multicolumn{4}{c|}{\bf DecomAtt} & \multicolumn{4}{c}{\bf BERT} \\ \cline{2-13}
& {\bf CLN} & {\bf PWWS} & {\bf GA-LM} & {\bf GA}
& {\bf CLN} & {\bf PWWS} & {\bf GA-LM} & {\bf GA}
& {\bf CLN} & {\bf PWWS} & {\bf GA-LM} & {\bf GA} \\ 

\hline

{\bf ORIG} & $\bf 80.4$ & $20.4$ & $38.3$ & $6.6$ & $\bf 81.9$ & $20.5$ & $39.2$ & $6.7$ & $\bf 90.5$ & $42.6$ & $56.7$ & $19.9$ \\

{\bf ADV} & $\bf 80.4$ & $67.9$ & $71.0$ & $59.5$ & $\bf 81.9$ & $71.7$ & $73.8$ & $65.2$ & $89.4$ & $68.2$ & $79.0$ & $58.2$ \\

{\bf IBP} & $79.3$ & $74.9$ & $75.0$ & $71.0$ & $77.3$ & $72.8$ & $73.7$ & $70.5$ & $--$ & $--$ & $--$ & $--$ \\

{\bf RAN} & $79.0$ & $65.7$ & $44.4$ & $27.8$ & $80.3$ & $67.2$ & $51.1$ & $30.6$ & $89.9$ & $\bf 72.7$ & $42.7$ & $27.0$ \\

{\bf DNE} & $79.8$ & $\bf 76.3$ & $\bf 75.3$ & $\bf 71.5$ & $80.2$ & $\bf 77.4$ & $\bf 76.7$ & $\bf 74.6$ & $89.3$ & $71.7$ & $\bf 80.0$ & $\bf 61.6$ \\

\hline
\hline

\end{tabular}
\end{center}

\end{table*}

\subsection{Effect of Parameters of Dirichlet Distribution}

Recall that the Dirichlet distribution is parameterized by a vector of $\boldsymbol{\alpha}$, and given a word $x_i$ different values of $\boldsymbol{\alpha}$ are used to control the degree in which its 1-hop and 2-hop neighbors to contribute to generate virtual adversarial examples, and also determines the size of the expansion from $\mathcal{C}(x_i)$ to $\mathcal{B}(x_i)$.
In order to reduce the impact on the clean accuracy, we let the expected weights of the 2-hop neighbors are $\lambda \in (0, 0.5]$ times of those of the (1-hop) nearest neighbors. We tried a few different values of $\alpha$ and $\lambda$ on IMDB to understand how the choice of them impact upon the performance. 
As shown in Table \ref{tb:alpha}, we found that if the value of $\alpha$ is fixed the greater the value of $\lambda$ the more robust the models will become, but the worse they perform on the clean input data. A small value of $\alpha$ seems to be preferable, which allows us to better simulate the discrete word substitution-based perturbations.

\begin{minipage}{\textwidth}
\begin{minipage}[t]{0.45\textwidth}
\setlength{\belowcaptionskip}{5pt}
\makeatletter\def\@captype{table}\makeatother\caption{Effect of Parameter $\boldsymbol{\alpha}$ on IMDB.}
\label{tb:alpha} 
\centering
\setlength{\tabcolsep}{0.9mm}
\begin{tabular}{l|cccc}

\hline
\hline

{$\bf \alpha$, $\bf \lambda$} & {\bf CLN} & {\bf PWWS} & {\bf GA-LM} & {\bf GA}  \\

\hline

{$0.1, 0.02$} & $\bf 86.2$ & $79.0$ & $76.0$ & $68.2$ \\

{$0.1, 0.1$} & $\bf 86.2$ & $81.4$ & $79.4$ & $75.4$ \\

{$0.1, 0.5$} & $84.8$ & $\bf 82.2$ & $79.8$ & $76.4$ \\

{$1.0, 0.02$} & $85.6$ & $78.8$ & $80.4$ & $75.6$ \\

{$1.0, 0.1$} & $85.1$ & $80.4$ & $\bf 80.8$ & $77.8$ \\

{$1.0, 0.5$} & $81.6$ & $78.6$ & $79.4$ & $\bf 78.2$ \\

\hline
\hline
\end{tabular}
\end{minipage}
\begin{minipage}[t]{0.54\textwidth}
\setlength{\belowcaptionskip}{3pt}
\makeatletter\def\@captype{table}\makeatother\caption{Ablation Study on IMDB.}
\label{tb:ablation} 
\centering
\setlength{\tabcolsep}{0.9mm}
\begin{tabular}{l|cccc}

\hline
\hline

{\bf Model} & {\bf CLN} & {\bf PWWS} & {\bf GA-LM} & {\bf GA}  \\

\hline

{\bf DNE} & $86.2$ & $81.4$ & $79.4$ & $75.4$ \\

{w/o EXPANSION} & $-0.1$ & $-14.2$ & $-24.0$ & $-45.0$ \\

{w/o ADV-TRAIN} & $+1.6$ & $- \, \, \, 7.8$ & $-19.8$ & $-34.6$ \\

{w/o COORD-UPD} & $-0.0$ & $- \, \, \, 4.2$ & $- \, \, \, 9.0$ & $-12.8$ \\

{w/o ENSEMBLE} & $-0.4$ & $- \, \, \, 1.8$ & $- \, \, \, 7.0$ & $- \, \, \, 9.4$ \\

\hline
\hline

\end{tabular}
\end{minipage}
\end{minipage}

\subsection{Ablation Study}

We conducted an ablation study on DNE over IMDB to analyze the robustness and generalization strength of different variants.
The ``w/o EXPANSION'' in the second row of Table \ref{tb:ablation} indicates that given any word $x_i$ in a sentence we generate virtual examples by sampling from $\mathcal{C}(x_i)$ instead of the expanded $\mathcal{B}(x_i)$ during the training.
The variant of DNE trained without using the adversarial training algorithm described in Section \ref{sec:adversarail} is indicated by ``w/o ADV-TRAIN''.
If the single-point update strategy is applied to train DNE, we still use the same gradient-guided optimization method to find adversarial examples over $\mathcal{B}(x_i)$, but the found adversarial example $\boldsymbol{x_j}$ is represented as $\boldsymbol{x_i} + \Delta$, where $\Delta$ is the distance between $\boldsymbol{x_i}$
and $\boldsymbol{x_j}$.
By such representation only $\boldsymbol{x_i}$ will be updated during the training instead of the embeddings of all its synonyms, and 
this variant is indicated by ``w/o COORD-UPD''. We also report in the last row the results predicted without using the ensemble method (i.e., $k = 1$) describe in Section \ref{sec:ensemble}. 

As we can see from Table \ref{tb:ablation}, the differences in accuracy among the variants of DNE are negligible on the clean data.
The key components to improve the robustness of the models in descending order by their importance are the following: sampling from the expanded convex hull $\mathcal{B}(x_i)$, combining with adversarial training, updating the word embeddings together, and using the ensemble to get the prediction.
We also observed that the stronger the attack method, the more effective these components.

\section{Conclusion}

In this study, we develop a novel defense algorithm to NLP models to substantially improve the robust accuracy without sacrificing their performance too much on clean data. 
This method is broadly applicable, generic,  scalable, and can be incorporated with negligible effort in any neural network.
A novel adversarial training algorithm is also proposed, which enables NLP models to defense against the strong attacks that search for the worst-case over all combinations of word substitutions.
We demonstrated through extensive experimentation that our adversarially trained smooth classifiers consistently outperform all existing empirical and certified defenses by a significant margin on IMDB, AGNEWS and SNLI across different network architectures, establishing state-of-the-art for the defenses against text adversarial attacks.




\small {
\bibliography{nips2020}
\bibliographystyle{plain} 
}




\section*{Appendix}

\subsection*{A.1 Experimental Details for Text Classification}

We report in Table \ref{tb:model_classification} and \ref{tb:hyper_classification} the values of hyperparameters used to train the text classification models, and the hyperparameter values of Dirichlet Neighborhood Ensemble (DNE) in Table \ref{tb:dne_classification}.
All models are trained on cross-entropy loss,
and their hyper-parameters are tuned on the validation sets. 

\begin{table*} [htbp] \small

\caption{\label{tb:model_classification} Hyperparameters for training the text classification models.}

\begin{center}
\begin{tabular}{l|cccc}

\hline
\hline

{\bf Model} & {\bf Word Embedding} & {\bf Hidden Size} & {\bf Layer} & {\bf Kernel Size}  \\

\hline

{\bf BOW} & $300$, GloVe \cite{emnlp-14:Pennington} & $100$ & $--$ & $--$ \\

{\bf CNN}  & $300$, GloVe \cite{emnlp-14:Pennington} & $100$ & $1$ & $3$ \\

{\bf LSTM} & $300$, GloVe \cite{emnlp-14:Pennington} & $100$ & $2$ & $--$ \\

\hline
\hline

\end{tabular}
\end{center}

\end{table*}

\begin{table*} [htbp] \small
\caption{\label{tb:hyper_classification} Training hyperparameters for the text classification (BOW, CNN, and LSTM) models. The same values were used for all training settings (plain, data augmentation, and robust training).}
\begin{center}
\begin{tabular}{l|c}
\hline
\hline

{\bf Hyperparameter} & {\bf Value} \\

\hline
{\bf Optimizer} &$0.5 \times 10^{-3}$, Adam \cite{iclr-15:Kingma} \\
{\bf Dropout} (word embedding) & $0.3$ \\
{\bf Weight decay} & $1 \times 10^{-4}$ \\
{\bf Batch size} & $32$ \\
{\bf Gradient clip}  & $(-1, 1)$ \\
{\bf Maximum number of epochs} & $20$ \\ 
\hline
\hline
\end{tabular}
\end{center}
\end{table*}

\begin{table*} [htbp] \small

\caption{\label{tb:dne_classification} Hyperparameters of DNE for text classification.}

\begin{center}
\begin{tabular}{l|c}

\hline
\hline

{\bf Hyperparameter} & {\bf Value} \\

\hline

{{\bf Dirichlet distribution parameter} $\alpha$ (the nearest neighbors)} & $0.1$ (\textbf{IMDB}), $1.0$ (\textbf{AGNEWS})  \\

{{\bf Parameter} $\lambda$ (neighbors of neighbors)}  & $0.1$ (\textbf{IMDB}), $0.5$ (\textbf{AGNEWS}) \\

{{\bf Step size} $\epsilon$ (adversarial training)} & $10$ \\

{{\bf Number of steps} (adversarial training)} & $3$ \\

{\bf Parameter} $r$ (ensemble method) & $3$ \\ 

\hline
\hline

\end{tabular}
\end{center}

\end{table*}

\subsection*{A.2 Experimental Details for Natural Language Inference}

All models take the pre-trained Glove word vectors as inputs and are trained on cross-entropy loss.
Their hyper-parameters are tuned on the validation sets.

\textbf{Bag of Words (BOW)}: We use a bag-of-word model with the same hyperparameters as shown in Table \ref{tb:model_classification} to encode the premise and hypothesis separately by summing their word vectors, then feeds the concatenation of these encodings to a two-layer feedforward network with a $300$-dimensional hidden state. We used the Adam optimizer (with a learning rate $0.5 \times 10^{-3}$), and set the dropout rate on word embedding to $0.3$, the weight decay to $1 \times 10^{-4}$, the batch size to $128$, the maximum number of epochs to $20$, and the gradient clip to $(-1, 1)$.

\textbf{Decomposable Attention (DecomAtt)}: We implemented the decomposable attention follows the original described in \cite{emnlp-16:Parikh} except for a few differences listed below:
\begin{itemize}[leftmargin=*]
\setlength{\itemsep}{0pt}
\setlength{\parsep}{0pt}
\setlength{\parskip}{0pt}
    \item We did not normalize GloVe vectors \cite{emnlp-14:Pennington} to have norm $1$.
    \item We used the Adam optimizer (with a learning rate of $0.5 \times 10^{-3}$) instead of AdaGrad.
    \item We used a dropout rate of $0.3$ on word embedding.
    \item We used a batch size of $128$ instead of 4.
    \item We clipped the value of gradients to be within $(-1, 1)$.
    \item We set the value of weight decay to $1 \times 10^{-4}$.
    \item We did not use the intra-sentence attention module.
\end{itemize}

\textbf{Bidirectional Encoder Representations from Transformers (BERT)}: We implemented the BERT follows the original described in \cite{naacl-18:Devlin} except for a few differences listed below:
\begin{itemize}[leftmargin=*]
\setlength{\itemsep}{0pt}
\setlength{\parsep}{0pt}
\setlength{\parskip}{0pt}
    \item We applied a ``bert-base-uncased'' architecture ($12$-layer, $768$-hidden, $12$-heads, $110$M parameters).
    \item We use the Adam optimizer (with a learning rate of $0.4 \times 10^{-4}$).
    \item We used a batch size of $8$.
    \item We set the number of epochs to $3$.
    \item We clipped the value of gradients to be within $(-1, 1)$.
    \item We set the value of weight decay to $1 \times 10^{-4}$.
    \item We used slanted triangular learning rates described in \cite{acl-18:Howard}, which first linearly increases the learning rate and then linearly decays it. 

\end{itemize}

We report in Table \ref{tb:dne_nli} the hyperparameter values of Dirichlet Neighborhood Ensemble (DNE) used for SNLI benchmark, and they are tuned on the validation set of SNLI. 

\begin{table*} [htbp] \small

\caption{\label{tb:dne_nli} Hyperparameters of DNE for natural language inference.}

\begin{center}
\begin{tabular}{l|c}

\hline
\hline

{\bf Hyperparameter} & {\bf Value} \\

\hline

{{\bf Dirichlet distribution parameter} $\alpha$ (the nearest neighbors)} & $1.0$ \\

{{\bf Parameter} $\lambda$ (neighbors of neighbors)}  & $0.5$ \\

{{\bf Step size} $\epsilon$ (adversarial training)} & $10$ \\

{{\bf Number of steps} (adversarial training)} & $3$ \\

{\bf Parameter} $r$ (ensemble method) & $3$ \\ 

\hline
\hline

\end{tabular}
\end{center}

\end{table*}

\end{document}